\documentclass[a4paper, 10pt, twocolumn]{article}  

\usepackage{amsfonts}
\usepackage{amsmath}
\usepackage{graphicx}
\usepackage{subfigure}
\usepackage{authblk}
\usepackage[utf8]{inputenc}

\RequirePackage{fix-cm}
\DeclareMathOperator{\pen}{pen}
\DeclareMathOperator{\crit}{crit}

\DeclareMathOperator{\st}{s.t.}

\renewcommand{\u}{\textbf{u}}
\renewcommand{\v}{\textbf{v}}
\newcommand{\w}{\textbf{w}}
\newcommand{\x}{\textbf{x}}

\graphicspath{{graphic/}}

\begin{document}

\title{An Empirical Comparison of $V$-fold Penalisation and Cross Validation for Model Selection in Distribution-Free Regression} 

%

\author[1, *]{Charanpal Dhanjal} 
\author[1]{Nicolas Baskiotis} 
\author[2]{St\'ephan Cl\'{e}men\c{c}on}
\author[1]{Nicolas Usunier}

\affil[1]{LIP6, UPMC, 4 Place Jussieu, 75252 Paris Cedex 05, France} 
\affil[2]{Institut Telecom, LTCI UMR Telecom ParisTech/CNRS No. 5141, 46 rue Barrault, 75634 Paris Cedex 13, France}

\date{\today}
\maketitle

\let\oldthefootnote\thefootnote
\renewcommand{\thefootnote}{\fnsymbol{footnote}}
\footnotetext[1]{Author for correspondence (charanpal.dhanjal@lip6.fr)}
\let\thefootnote\oldthefootnote

\begin{abstract}
Model selection is a crucial issue in machine-learning and a wide variety of penalisation methods (with possibly data dependent complexity penalties) have recently been introduced for this purpose. However their empirical performance is generally not well documented in the literature. It is the goal of this paper to investigate to which extent such recent techniques can be successfully used for the tuning of both the regularisation and kernel parameters in support vector regression (SVR) and the complexity measure in regression trees (CART). This task is traditionally solved via $V$-fold cross-validation (VFCV), which gives efficient results for a reasonable computational cost. A disadvantage however of VFCV is that the procedure is known to provide an asymptotically suboptimal risk estimate as the number of examples tends to infinity. Recently, a penalisation procedure called $V$-fold penalisation has been proposed to improve on VFCV, supported by theoretical arguments. Here we report on an extensive set of experiments comparing $V$-fold penalisation and VFCV for SVR/CART calibration on several benchmark datasets. We highlight cases in which VFCV and $V$-fold penalisation provide poor estimates of the risk respectively and introduce a modified penalisation technique to reduce the estimation error. 

\end{abstract}

\section{Introduction}

Learning algorithms generally depend on a small number of real-valued or discrete parameters such as the size of a tree in hierarchical methods, the stopping criteria in boosting algorithms or explicit regularisation/smoothing parameters. These parameters naturally determine the complexity of the output function, and by doing so, also strongly influence generalisation ability. In a general sense the more "complex'' the learnt function is, the more likely it is to overfit to the data. On the contrary, a simple predictor will be suboptimal if the data is informative with regard to the learning problem. From the model selection point of view the challenge consists in selecting values of the parameters of interest with a theoretical risk as small as possible.
From a global perspective, there exist essentially two major approaches to model selection: methods related to data-splitting, with cross-validation \cite{Allen:74} and its variants, and methods related to penalisation of the empirical risk (that obtained on the training set), with in particular the Structural Risk Minimisation principle \cite{VapCher:74}. Penalisation-based approaches aim to approximate the ideal model by adding a penalty or complexity-based term to the empirical risk, generally based on theoretical arguments (\textit{i.e.} on probabilistic distribution-free upper bounds for the excess of risk). $V$-fold cross-validation (VFCV in abbreviated from) is widely used in machine-learning practice due to its (relative) computational tractability and empirical evidence of its good behaviour. However, there is little theoretical justification concerning VFCV \cite{ArlotCelisse:10} and the question of an automatic choice of the parameter $V$ remains widely open \cite{BreiSpec:92}. On the other hand, 
penalisation procedures generally suffer from two possible drawbacks, despite the fact that they are theoretically well-
founded: either they are designed in a simplified framework and thus are not robust to complex situations (typically the case for Mallows' $ C_{p}$ in regression \cite{Mallows:73}), or they are of general purpose, as for instance the so-called Rademacher complexities \cite{bartlett03rademacher}, but are inaccurate in many cases. In order to refine Rademacher penalties, several authors proposed localisation techniques, giving rise to local Rademacher complexities \cite{Kolt:06}, but these more accurate capacity functions are essentially of theoretical interest and cannot be used in practice due to the presence of unknown constants in their definition.

Combining both the robustness of cross-validation estimates and theoretical guarantees of penalisation procedures, a new type of general purpose penalisation procedures, called $V$-fold penalisation, has been recently proposed \cite{Arl:2008a}. Both empirical and mathematical evidence of its efficiency have been shown in a heteroscedastic with random design regression framework, when considering the selection of finite-dimensional histogram models.
While the selection of regressograms studied in \cite{Arl:2008a} is convenient for theory since it allows precise mathematical investigations and is however general enough to show some relevant complex phenomena, we investigate in this paper the behaviour of $V$-fold penalties, and compare it to VFCV, for the tuning of the hyperparameters involved in the Support Vector Regression algorithm (SVR, \cite{drucker97svr}) and Classification and Regression Trees (CART, \cite{breiman1984classification}) for regression. Indeed, these algorithms are two of the most extensively used regression tools in a wide variety of areas and the choice of efficient hyperparameters is known to be a decisive step of the learning process to attain good generalisation performance. Model selection for SVR has been addressed by several authors and many attempts, theoretically well-founded, have been proposed to answer this problem, among which: estimation of the hyperparameters from the data and the level of noise \cite{Smola98}\cite{Kwok01}\cite{Cherka04}, leave-one-out bounds for SVR \cite{ChangBounds05}.
However,  methods based on resampling procedures for the evaluation of the risk of each model have been proven to be significantly better than most of the other proposed automatic procedures \cite{Cherka04,Ong10} and VFCV is generally the chosen method in practice \cite{SmolaTuto04}. For CART regression, the issue of model selection has not received as much attention, however \cite{gey2005model} provides a theoretical validation of the standard CART pruning criterion. 
In this paper, our aim is to study $V$-fold penalisation for model selection and give insights into situations when one can improve on VFCV in practice. Particularly, the comparison of $V$-fold penalisation with VFCV on the problem of SVR and CART calibration takes importance, due to the highlighted relevance of VFCV in this central issue. 

The remainder of this paper is organised as follows: Section \ref {section_framework} describes the statistical framework related to model selection for kernel SVR and CART. In Section \ref {section_Vfold_cross_val} we recall VFCV and related works, we introduce in Section \ref{section_Vfold_pen} $V$-fold penalisation and our improved penalisation approach. Experiments are addressed in Section \ref {section_experiments}, and conclusions are presented in Section \ref{section_conclu}.

\section{Background and Preliminaries\label{section_framework}}

As a first go, we outline the statistical setting of the model selection we shall subsequently study (generally referred to as the \textit{distribution-free regression} setup). Here and throughout, a column vector is written in bold lowercase \textit{e.g.} $\x$. Let $\mathcal{X\times Y}$ be a measurable space endowed with an unknown probability measure $P$, with $\mathcal{Y} = \mathbb{R} $. $\mathcal{X}$ is called the \emph{input space} and is usually a compact subset of $\mathbb{R} ^{d}$, $d\geq 1$, and $\mathcal{Y}$ is the \emph{target space}. We observe $n$ i.i.d. labelled observations or examples $S = \{ \left( \x_{1},y_{1}\right) ,\ldots,\left( \x_{n},y_{n}\right) \} \subset \mathcal{ X\times Y}$. Furthermore, $\left(X,Y\right) $ denotes a generic random variable, independent from the data $S$, drawn from $P$. Let $\mathcal{S}$ be the set of all measurable functions $s: \mathcal{X} \rightarrow \mathcal{Y}$ mapping from the input to target space.  In the present paper, focus is on the mean absolute 
deviation: 
\begin{align*} 
L(s) = \mathbb{E}[\vert s(X) - Y\vert].
\end{align*}
The regression task can thus be rewritten in these notations as finding minimum of the so-called least absolute Bayes loss $s_{\ast }$, defined by:
\begin{displaymath}
L(s^*)= \min_{s:\mathcal{X\rightarrow Y}} L(s). 
\end{displaymath} 

\subsection{Support Vector Regression} 

The SVR prediction function is of the form $f(\x) = \langle \w, \x \rangle + b$ where $\w \in \mathbb{R}^d$ is a weight vector and $b$ is a constant. In this case, one is interested in errors greater than a user-defined value $\epsilon \in \mathbb{R}^+$ (known as the $\epsilon$-insensitive loss). Hence the optimisation task can be written as: 
\begin{equation}
\begin{array}{r l} 
t = \arg \min_{\w, b} &  \frac{1}{2} \|\w\|^2 + C \sum_{i=1}^n (\xi_i + \xi_i^*) \\
\st & y_i - \langle \w, \x_i \rangle - b \leq \epsilon + \xi_i \\
& \langle \w, \x_i \rangle - y_i + b \leq \epsilon + \xi_i^* \\ 
& \xi_i, \xi_i^* \geq 0,  
\end{array} \label{eqn:svropt} 
\end{equation}
where $\xi_i$ and $\xi_i^*$ are slack variables, and $C$ is a \emph{user-defined} trade-off between minimising the norm of the weight vector $\w$ (which can be seen as  regularisation) and penalising errors greater than $\epsilon$. A high value of $C$ thus corresponds to a low regularisation level and the objective becomes then closer to that of minimising the empirical risk. The value of $\epsilon$ affects the number of Support Vectors (SV's in short), with larger values resulting in fewer SV's. In a slight abuse of notation, minimum values of $\w$ and $b$ form a prediction function $t$. 
The SVR algorithm is often performed using kernels to model non-linear functions, where a kernel function $\kappa:\mathcal{X\times }\mathcal{X\rightarrow }\mathbb{R}$ is used to find the inner product of the transformation of the input space $\mathcal{X}$ into its associated Reproducing Kernel Hilbert Space (RKHS), denoted by $\mathcal{H}_{\kappa}$. Note that $\kappa$ can be written in terms of a transformation $\phi$ from input to kernel space $\left\langle \u, \v \right\rangle_{\mathcal{H}_{\kappa}} = \left\langle \phi(\u), \phi(\v) \right\rangle = \kappa(\u, \v)$.   
Kernels functions usually depend on one or a few hyperparameters, \textit{e.g.} polynomial, Gaussian Radial Basis Function (RBF) and sigmoid kernels \cite{ScholSmola:02}. The Gaussian RBF kernel is one of the most commonly used kernels and Boser, Guyon and Vapnik suggested its widespread use \cite{boser92svm}\cite{GuyonBoserVapnik:93}\cite{Vapnik:95}. In the experiments described in Section \ref{section_experiments}, we therefore consider the Gaussian RBF kernel,  
\begin{displaymath}
  \kappa_{\gamma }( \x,\x^{\prime }) =\exp( -\gamma \left\Vert \x-\x^{\prime }\right\Vert ^{2}),  \label{Gaussian_kernel}
\end{displaymath}
\nolinebreak
which depends on one real-valued positive parameter $\gamma$. In the following we denote the Gaussian RBF kernel $\kappa_{\gamma }$, $\gamma \in \mathbb{R}_{+}$. 
The optimal value of the regularisation parameter $C$ can significantly change, and depends on the data. To ensure the performance in prediction of the SVR algorithm, the regularisation parameter, as well as the kernel, should thus be calibrated in each application. Formally, the question to be addressed is to find the best parameters $\left( \gamma ,C, \epsilon \right) $ in terms of prediction. We thus aim at estimating the  \emph{oracle}, which is the model with the smallest risk, (where $t\left( \gamma ,C, \epsilon \right)$ is the SVR learnt using parameters $\gamma ,C, \epsilon $), 
\begin{displaymath} 
 \arg \min_{\left( \gamma,C, \epsilon\right) \in \mathbb{R}_{+}^3} L\left( t\left( \gamma ,C, \epsilon \right) \right),
\end{displaymath}
\nolinebreak
which is unknown since it depends on the law $P$ of data and which optimises the least squares error.  

\subsection{CART Regression} 

\begin{figure}[ht]
\centering
\includegraphics[height=50mm]{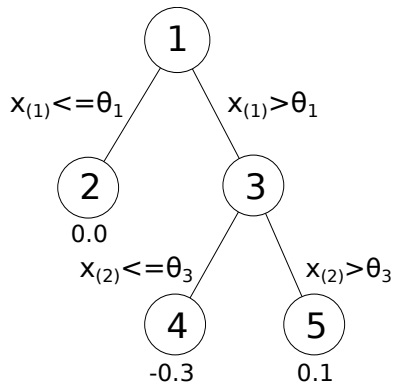} 
\caption{\label{fig:tree} An example of a decision tree.} 
\end{figure}

Another important algorithm for regression is CART  in which one learns a tree like the one exemplified in Figure \ref{fig:tree}. To regress a new example $\x$, it is filtered down to a leaf node via a decision at each link and then assigned a real number target. In the illustration, if the first feature of $\x$, $\x_{(1)} \leq \theta_1$, for some threshold $\theta_1$, then it is labelled $0.0$. Similarly, if $\x_{(1)} > \theta_1$ and $\x_{(2)} > \theta_3$ then the example is labelled $-0.3$. Regression trees have been successfully used in a variety of applications in such as vector quantisation \cite{chou1989optimal}, meteorology \cite{burrows1997cart}  and medicine \cite{wernecke1998validating} for example, and have the crucial advantage of being easy to interpret and easy to compute.    
To construct a regression tree one starts with the root node which contains all of the training examples $S_0 = S$. One then decides how to split the examples based on a feature $k$ and a threshold $\theta$. Given a choice of these values, the left child contains the examples $S_L = \{(\x, y) \in S_0 \; | \; \x_{(k)} \leq \theta \}$ and the right one is $S_R = \{(\x, y) \in S_0 \; | \; \x_{(k)} > \theta \}$. The optimal feature-threshold pair $k^*, \theta^*$ is found by minimising the squared error of the split, i.e. $k^*, \theta^*$  are found using: 
\begin{displaymath} 
\begin{array} {r l}
\arg  \underset{k, \theta}{\min} & \sum\limits_{(\x, y) \in S_L} (y - \mu_{S_L})^2 + \sum\limits_{(\x, y) \in S_R} (y - \mu_{S_R})^2   \\ 
\st & S_L = \{(\x, y) \in S_0 \; | \; \x_{(k)} \leq \theta \} \\
& S_R = \{(\x, y) \in S_0 \; | \; \x_{(k)} > \theta \}, 
\end{array}
\end{displaymath}
where $\mu_{S_L} = \frac{1}{|S_L|}\sum_{(\x, y) \in S_L} y$ and $\mu_{S_R} = \frac{1}{|S_R|}\sum_{(\x, y) \in S_R} y$ are the mean labels for the left and right nodes and hence $(y - \mu_{S_0} )^2$ is the squared error between $y$ and the mean label. A simple way to solve this optimisation is to iterate through each feature and threshold and choose the one with the lowest objective value. After splitting on the root node, one recursively splits on the resulting child nodes until no more splits are possible, \textit{i.e.} a node contains fewer examples than a user defined value $\ell$.  
Following the growing phase, one prunes the resulting tree as smaller trees have been shown to improve generalisation error. In CART, one uses an approach called \emph{cost complexity pruning} which generates a series of trees pruned from the original tree and then selects one of the trees in the sequence. For the $i$th node in the unpruned tree which contains examples $S_i$, one computes the error if the tree was pruned at that node and compares it to the error if the subtree starting at that node $R_i$ is kept. The difference in these errors divided by the number of leaves of the subtree gives an indication of the error difference per leaf, i.e. 
\begin{displaymath} 
 \alpha_i = \frac{\tilde{L}(S_i, \hat{R_i}) - \tilde{L}(S_i, R_i)}{|R_i|_l - |\hat{R_i}|_l},
\end{displaymath}
in which $\tilde{L}(S_i, R_i)$ is the squared error of a set of examples $S_i$ using subtree $R_i$, $\hat{R_i}$ is the root of $R_i$, and $|R_i|_l$ is the number of leaves in $R_i$. In simple terms, the higher the value of $\alpha_i$, the bigger the reduction in error of the subtree per leaf. One can compute $\alpha_i$ for all nodes in the tree and hence, if we prune nodes with $\alpha_i$ greater than a threshold $\sigma \in \{\alpha_1, \ldots \alpha_{|T|}\}$, where $|T|$ is the number of nodes in the tree, we obtain a sequence of trees which decrease in size as $\sigma$ increases. 
Instead of choosing $\sigma$ directly in the model selection stage, we pick a threshold $t$ and choose the largest tree smaller or equal in size to $t$. Therefore, as before, the model selection task can be written in terms of search for the parameter $t$. We aim at estimating the  oracle (where $f\left( t \right)$ is the decision tree learnt using parameter $t$), 
\begin{displaymath} 
 \arg \min_{ t \in  \mathbb{Z}} L\left( f\left( t \right) \right),
\end{displaymath}
Estimating the oracle is a model selection task, each model being represented here by a fixed value of $t $, where penalisation is a natural way to proceed, as explained in Section \ref {section_Vfold_pen} below. However, let us first briefly recall the method which is usually employed for model selection, namely $V$-fold cross-validation.

\section{$V$-fold Cross Validation\label{section_Vfold_cross_val}}

The idea of cross-validation for model selection is to estimate the risk of the considered estimator on each model by using a repeated data-splitting scheme, and then to select the model that minimises these estimates of the risk. The fact that data-splitting strategies give accurate estimates of the risks only relies on the independence between each training and testing set. Consequently, the interest of CV is that it is based on a heuristic that can be applied with great universality.
Many data-splitting rules have been proposed, such as leave-one-out (LOO, \cite{Allen:74}), leave-p-out (LPO, \cite{Shao:93}), balanced incomplete CV (BICV, \cite{Shao:93}), repeated learning-testing (RLT, \cite{BreiFriedOlshStone:84}). $V$-fold cross-validation (VFCV) was introduced by Geisser \cite{Geisser:75}, see also  \cite{BreiFriedOlshStone:84} as a computationally efficient alternative to LOO cross validation. We will consider primarily VFCV, which is certainly the most commonly used cross-validation rule in practice. Moreover, it is generally the procedure which is considered for the calibration of the SVR hyperparameters \cite{SmolaTuto04} \cite{HastieTibFried:01}.
In VFCV, the examples are partitioned into $V$ subsamples of $n/V$ examples each (with a maximal deviation of one) $B_1, \ldots, B_V$. At the $j$th fold one trains on $S \setminus B_j$  and then evaluates the error on $B_j$, and one averages the errors over all $V$ folds. For model selection this is repeated over a grid of parameters in order to select those with the lowest error. 
Despite the generality of the heuristic underlying the procedure, there are two drawbacks in the $V$-fold cross-validation method for model selection. First, at a fixed $V$, the procedure is known to be asymptotically suboptimal in the sense that the risk of the selected estimator is not equivalent to the risk of the oracle when the number of data tends to infinity. More precisely, \cite{Arl:2008a} shows in a heteroscedastic with random design regression framework, that VFCV with fixed $V$ satisfies an oracle inequality with a constant $A>1$ which relates the excess risk of the selected model to the excess risk of the oracle, and that this suboptimal constant cannot be improved asymptotically. The keystone of such a result is that at a fixed $V$, the VFCV criterion is biased compared to the true risk \cite{Burman:89}\cite{rodriguez2010sensitivity}. Indeed, since the validation sets are independent of their respective training sets, the expectation of the VFCV criterion can be related to the expectation of 
the true risk $\mathbb{E}[\crit_{\text{VFCV}}s\left(Q\right)]$ for a learner $s$ with parameter set $Q \in \mathcal{Q}$ as follows, 
\begin{eqnarray*}
\mathbb{E}\left[ L_S^{(j)}\left( s^{(-j)}\left( Q  \right) \right)  \right] = \mathbb{E}\left[ L\left(s^{[-1]}\left( Q  \right) \right) \right] \text{ ,}
\end{eqnarray*}
where $s^{[-1]}\left( Q  \right) $ is the output of the learner trained with $\left( 1-1/V\right) n$ i.i.d. examples, $s^{(-j)}$ is the learner trained with $S \setminus B_j$, and $L_S^{(j)}$ is the loss with respect to the partition $B_j$ of $S$. Since the true risk generally decreases with more data, it appears that the expectation of VFCV criterion roughly overestimates the expectation of the true risk, and that this bias should be decreasing whenever $V$ increases.
The previous observation suggests that, in order to mimic the oracle in terms of performance in prediction, one should take a $V$ which is as large as possible. This is where appears the second drawback concerning VFCV: there is no rule in practice to choose the optimal $V$. Indeed, the best CV estimator of the risk is not necessarily the best model selection procedure, and \cite{BreiSpec:92} highlight that LOO is the best estimator of the risk, whereas $10$-fold cross-validation is more efficient for model selection purpose. This can be explained by the fact the bias in the VFCV estimation of the risk is actually an advantage for model selection with a few or medium number of examples, contrary to the asymptotic framework. Indeed, as claimed in \cite{Arl:2008a} a slightly over-pessimistic estimation of the risk, as in VFCV, gives for a fixed number of observations a more robust model selection procedure and roughly contradicts the bad effects of the variance of risk estimation.

\section{$V$-fold Penalisation\label{section_Vfold_pen}}

Penalisation is a natural strategy for the task of estimating the oracle $s\left( Q ^{\ast } \right) $. Indeed, the definition of the oracle can be rewritten as the sum of the empirical loss and an unknown term, which is thus an ideal penalty in the sense that it allows one to recover the oracle: 
\begin{equation*}
\arg \min_{Q  \in \mathcal{Q}}   L_S\left(s\left( Q \right) \right)  + \pen_{\text{id}}\left( Q  \right),
\end{equation*}
in which $s(Q)$ is a function mapping from input to target space under hyperparameters $Q$, and the ideal penalty is as follows, 
\begin{displaymath}
 \pen_{\text{id}}\left( Q  \right) = L\left(  s\left( Q  \right) \right) -L_S\left( s\left( Q \right) \right).
\end{displaymath}
\nolinebreak
Hence, penalisation aims at mimicking the oracle by selecting, for a known penalty function the estimator 
\begin{displaymath} 
\arg \min_{Q \ \in \mathcal{Q}}  L_S\left(s\left( Q  \right) \right)  + \pen \left( Q  \right). 
\end{displaymath}
\nolinebreak 
A good penalty in terms of prediction is one which gives an accurate estimate of the ideal penalty $\pen_{\text{id}}$.
The central idea of $V$-fold penalties proposed in \cite {Arl:2008a} is to directly estimate the ideal penalty by a subsampled version of it. For some constant $C_{V}\geq V-1$, the $V$-fold penalty $\pen_{\text{VF}}\left( Q \right)$ is
\begin{eqnarray*} 
 \frac{C_{V}}{V}\sum_{j=1}^{V}  \left[ L_S( s^{(-j)}( Q) ) 
- L_S^{(-j)}(s^{(-j)}( Q))\right], \label{penVF}
\end{eqnarray*} 
and so the corresponding selected hyperparameters are given by 
\begin{displaymath}
 \arg \min_{Q \in \mathcal{D}}  L_S\left( s\left( Q \right) \right) +\pen_{\text{VF}}\left( Q  \right),
\end{displaymath}
\nolinebreak
where $\mathcal{D}$ is a discrete grid upon the set $\mathcal {Q}$. The $V$-fold penalty indeed mimics the structure of the ideal penalty, in such a way that the quantities related to the unknown law of data $P$ (respectively to the empirical measure $P_S$) are replaced by quantities related to the empirical measure $ P_S$ (respectively to the subsampling measures $P_S^{(-j)}$), in the same analytic manner. The design of the $V$-fold penalties is thus an adaptation of Efron's resampling heuristics \cite{Efron:79} to the subsampling scheme of the $V$-fold procedure.
It has been shown in \cite{Arl:2008a} by considering the selection of regressograms in a heteroscedastic regression framework, that $V$-fold penalisation with $C_{V}=V-1$ is asymptotically optimal for a fixed $V$, whereas in this case, VFCV is asymptotically suboptimal, due to its bias on the estimation of the risk. Moreover, the choice of $C_{V}=V-1$ in the definition of the $V$-fold penalty corresponds to the Burman's corrected VFCV criterion \cite{Burman:89}. Therefore, we use in our experiments $C_{V}=V-1$ although we also explore different values. 
Another advantage, highlighted in \cite{Arl:2008a}, of $V$-fold penalisation, compared to VFCV, is that it seems to be more regular with respect to the choice of $V$. At a heuristic level, this can be explained by observing that since $V$-fold penalisation corrects the bias of VFCV, it is only variability of $V$-fold estimates that matters here, a variability which is smaller for a larger $V$.
Finally, it should be noted that the constant $C_{V}$ in the definition of the $V$-fold penalty can be viewed as a degree of freedom, which potentially allows to deal with the bias of the proposed risk estimation, without varying the value of $V$, contrary to VFCV where only $V$ fixes simultaneously and in a tricky manner, the bias and variance of the risk estimation. In \cite{Arl:2008a} it is shown that choosing $C_V$ to overpenalise (\textit{i.e.} $\pen_{\text{VF}}$ is larger than $\pen_{\text{id}}$ even in expectation) can improve prediction performance when the signal to noise ratio is small. The choice is a difficult one however, and according to empirical results on synthetic datasets, it depends on the sample size, noise level and smoothness of the regression function. 

\section{A Complexity-Based Selection of $C_V$}\label{sec:complexity}

In practice, as we shall later see, for a fixed training set and $V$, the approximate penalty as given by $\pen_{\text{VF}}$ often poorly approximates the ideal penalty and it cannot be improved by varying the penalisation constant $C_V$. To study the cause of the problem we analyse the $\pen_{\text{VF}}$ criterion relative to the ideal penalty. Consider the first term in the sum of $\pen_{\text{VF}}$, $ L_S\left( s^{(-j)}\left( Q  \right) \right)$, and write it in terms of the loss on the training and test set: 
\begin{eqnarray*} 
& &  \frac{1}{n} \sum_{i \in S^{(-j)}} \ell_i(s^{(-j)}(Q)) +  \frac{1}{n} \sum_{i \in S^{(j)}}\ell_i(s^{(-j)}(Q)) \\ 
&=&  \frac{V-1}{V} L_S^{(-j)}(s^{(-j)}(Q) + \frac{1}{V} L_S^{(j)}(s^{(-j)}(Q),
\end{eqnarray*}
in which $\ell_i(\cdot)$ is the loss for the $i$th example. The link between the lines can be seen by noting that 
\begin{displaymath} 
L_S^{(-j)}(s^{(-j)}(Q) = \frac{V}{(V-1)n} \sum_{i \in S^{(-j)}} \ell_i(s^{(-j)}(Q)).
\end{displaymath} 
When we put the above form of $ L_S\left( s^{(-j)}\left( Q  \right) \right)$ into $\pen_{\text{VF}}$ we obtain
\small
\begin{displaymath}
 \frac{C_V}{V} \left[\frac{1}{V} \sum_{j=1}^V \left( L_S^{(j)}\left( s^{(-j)}\left( Q  \right) \right)  -L_S^{(-j)}\left( s^{(-j)}\left( Q  \right) \right)  \right) \right], 
\end{displaymath}
\normalsize
and the term inside the square brackets is the empirical expectation of the error on the test set minus the error on the training set. One can say that this an approximation of the ideal penalty using $(V-1)n/V$ examples since the loss term on the right side is computed over $S \setminus B_j$. A variety of error bounds have the penalty proportional to a complexity measure and inversely proportional to the number of examples to some power of a learning rate $\beta(Q)$ (see \cite{liang2010interaction} for example). In other words, for a learner $s$ with parameters $Q$ we consider the following form of the penalty:  
\begin{equation} \label{eqn:penV}
 \pen_V(Q) = \frac{C_V}{V} \frac{D(Q)}{(n(V-1)/V)^{\beta(Q)}} 
\end{equation}
where $D(Q)$ is the complexity of $s(Q)$ and $\beta(Q)$ is a learning rate, and we have replaced the square bracketed term in $\pen_{\text{VF}}$ with $D(Q)/(n(V-1)/V)^{\beta(Q)}$. A learning rate of 0 implies a large penalty and that we have overfitted the data and hence, for a fixed $V$ and sample size we do not learn anything (in other words one predicts on a test set randomly). As the sample size increases one continues to overfit and hence the penalty term is $C_V D(Q)/V $ regardless of the sample size. In contrast when $\beta(Q) = 1$ the penalty is small, and rapidly decreases with the sample size, and also with $V$. A limit of 1 for $\beta(Q)$ is natural for the learning rate since this is the bound often used in complexity bounds. 
The ideal penalty has the form $D(Q)/n^{\beta(Q)}$ for $n$ examples and hence we would like to choose $C_V$ above so that 
\begin{equation} \label{eqn:penVIdeal}
\frac{C_V}{V} \frac{D(Q)}{(n(V-1)/V)^{\beta(Q)}} = \frac{D(Q)}{n^{\beta(Q)}},
\end{equation} 
and solving gives  $C_V = (V-1)^{\beta(Q)}/V^{\beta(Q)-1}$. A learning rate of $0$ which occurs with complex models (e.g. large decision trees) implies $C_V = V$ and similarly for small models where $\beta(Q)=1$ we have $C_V = V-1$ as suggested asymptotically above. In this latter case we recover exactly the value of $C_V$ suggested in \cite{Arl:2008a}. On the whole, $\pen_V(Q)$ is an estimation of the ideal penalty using $\pen_{\text{VF}}$ and the model complexity $D(Q)$.   
It remains to consider how one computes the learning rate. We equate Eq. \eqref{eqn:penV} with $\pen_{\text{VF}}$ and then taking logs results in $\log(\pen_{VF}(Q)/C_V)$ equal to  
\begin{displaymath} 
-\log(V) - \beta(Q) \log(n(V-1)/V)  + \log(D(Q)). 
\end{displaymath}
One finds the gradient of $\log(\pen_{VF}(Q)/C_V) + \log(V)$ versus $\log(n(V-1)/V)$ for a selection of different $V$ values whilst fixing $Q$ and $n$, in order to find the learning rate $\beta(Q)$.   

\section{Experimental Setup \label{section_experiments}}

We study the behaviour of VFCV and $V$-fold penalisation on a collection of benchmark datasets. The scikit-learn library in Python  \cite{scikit-learn} is used to generate the output of the RBF SVR and CART algorithms. 

In total, 10 datasets from the UCI machine learning repository \cite{FrankAsuncion2010} and DELVE \cite{Neal98assessingrelevance} are used. Each dataset is split into 100 training and test realisations/sets after being processed so that the examples and labels have zero mean and unit standard deviation. Details are provided in Table \ref{tab:datasetInfo}. When comparing model selection algorithms, a statistically significant improvement of one method over another is such that the mean error is greater and by using a paired $t$-test. For the $t$-test we take the sample of errors over all realisations for 2 methods, then compute a $p$ value and reject the null hypothesis, that the means are equal, if $p < 0.1$. In all experiments we use the mean absolute error, i.e. for a prediction function $f: \mathcal{X} \rightarrow \mathcal{Y}$, the error is $\frac{1}{n} \sum_{i=1}^n \|f(x_i) - y_i \|_1$. 

\begin{table}
\centering

\begin{center}
 \begin{tabular}{l l l l l} 
\hline 
Dataset & Learn & Test & $d$ & Abrv.  \\ 
\hline 
abalone & 835 & 3342 & 8 & ab\\
add10 & 2937 & 6855 & 10 & ad\\
comp-activ & 2457 & 5735 & 22 & ca\\
concrete & 309 & 721 & 8 & cc\\
parkinsons-motor & 2937 & 2938 & 20 & pm\\
parkinsons-total & 2937 & 2938 & 20 & pt\\
pumadyn-32nh & 3276 & 4916 & 32 & pd\\
slice-loc & 26750 & 26750 & 385 & sl\\
winequality-red & 1066 & 533 & 11 & wr\\
winequality-white & 3265 & 1633 & 11 & ww\\
\hline 
 \end{tabular}
\end{center}
\caption{Information on the benchmark datasets used. There are 100 learn/test splits for each dataset. }
\label{tab:datasetInfo}
\end{table}

\subsection{Model Selection}

In all of the following experiments we use a grid to approximate the set of hyperparameters. The SVR penalty is chosen as $C \in \{2^{-10}, 2^{-8}, \ldots, 2^{12} \}$, the kernel width as $\gamma \in \{2^{-10}, 2^{-8}, \ldots, 2^2 \}$, and $\epsilon \in \{2^{-4}, 2^{-3}\}$. More sophisticated ways of searching in the hyperparameter space actually exist, such as the iterative process derived from the so-called active sets method and used in \cite{HastieRossetTibshirani:04,Cherka04,Ong10} to walk along the entire path of the SVR: $\gamma $ is fixed and \textit{all} values of $C$ are considered. Others heuristics involve e.g.  genetic algorithms \cite{He08}, local search methods \cite{Momma02}. However, some degeneracies can occur and so, a search on a grid should be more stable.  Moreover, the grid-search has also the advantage of being easily parallelised, because each value of $\left( \gamma ,C, \epsilon \right)$ is independent from the others. In the case of CART regression, we pick the bound on the tree 
size $t$ from $\{2^1-1, \mbox{round}(2^{1.5}-1), \ldots, 2^7-1\}$. 

An important characterisation of the model picked during the selection phase is its complexity. In all of the model selection techniques we choose a set of parameters over $n(V-1)/V$ examples however the final predictor is training using all $n$ examples. Ideally, we would like the complexity to be identical in both model selection and whilst training using all examples, since the penalty is a function of complexity (Equation \ref{eqn:penV}). For the SVR the norm of the weight vector is the measure of complexity used in error bounds (see \cite{SmolaTuto04}). For this reason, we compute the mean norm of the SVR weight vector $\|\w\|$, for each value of $C, \gamma, \epsilon$ used in model selection and store $\|\w^*\|$, the norm corresponding to the lowest error, as well as the corresponding values $\gamma^*, \epsilon^*$. When we train using all $n$ examples, we again compute $\|\w\|$'s corresponding to each value of $C$, for $\gamma^*, \epsilon^*$, and choose $C$ with corresponding norm closest to $\|\w^*\|$. 
In the case of CART trees, we can be slightly more direct: for the optimal bound on the tree size, $t^*$, we compute the real mean tree size found during model selection and round to the  nearest integer  $\hat{t}$. The value of $\hat{t}$ is then used to train over all $n$ examples. 

\subsection{Primary Setup}

In order to test the model selection techniques we take random training subsamples of either $50$, $100$ or $200$ examples of the learning sets to observe model selection on a limited number of examples. Furthermore, we test using $2, 4, \ldots, 12$ folds. Model selection is performed using each subsample and then  SVR or CART is trained using the optimal parameters and the entire subsample. This is repeated for each realisation and results are averaged over the entire set of realisations. As well as recording the error obtained using model selection over the realisations, we also store the difference between the ``ideal'' and approximated penalty. This former quantity is computed simply as the difference between the $V$-fold penalty and the penalty as computed using the test set. All results for $V$-fold penalisation are evaluated with $C_V = (V-1) \alpha$ with $\alpha \in  \{0.6, 1.2, \ldots, 1.6\}$ being the multiplier for the penalisation. We denote the types of model selection  methods as: VFCV, $V$-
fold penalisation (\texttt{PenVF}) and $V$-fold penalisation using a learning rate (\texttt{PenVF+}).

For the \texttt{PenVF+} method one needs to compute learning rates for each model (set of parameters). We use the same training sets as above and vary $V$ from the set $\{2, 3, \ldots, 12\}$. The quantities  $\log(\pen_V(Q)) + \log(V)$ versus $\log(n(V-1)/V)$ are computed and the gradient, found using linear regression, provides $\beta(Q)$ which in turn is used to calculate $C_V = (V-1)^{\beta(Q)}/V^{\beta(Q)-1}$. As this estimation of $\beta(Q)$ can be unstable especially with small training sets we clip its value to lie within the valid range $[0, 1]$. 

\section{Experimental Results} 

\subsection{Comparison of Penalisation and VFCV with $V=2$}  

We start by studying the SVR results in Table \ref{tab:allDatasets} which shows errors for all datasets when $V=2$, $\alpha=1.0$. We consider $V=2$ in this case since it provides the greatest distinction between the model selection methods. For \texttt{PenVF+} many of the results are comparable to VFCV when we also consider the standard deviations. As the same time, \texttt{PenVF+} does not always improve upon \texttt{PenVF}. In contrast \texttt{PenVF} can perform significantly worse than both VFCV and \texttt{PenVF+}, for example with  \texttt{abalone}, \texttt{winequality-red} and \linebreak \texttt{winequality-white}. Also of note is that the difference in error between VFCV and \texttt{PenVF} does not improve with $m=200$ with  \texttt{abalone} for example: it is $0.08, 0.089, 0.089$ with $m=50, 100, 200$. 

Also shown at the bottom of Table \ref{tab:allDatasets} is the equivalent CART results. It is evident that error rates are generally worse than the SVR with the exception of \texttt{pumadyn-32nh}. Also, we see that penalisation provides a larger advantage relative to VFCV in this case. One explanation is that CART is more sensitive to its hyperparameters. We observe that \texttt{PenVF+} is equivalent or improves over VFCV in nearly every case, and there are $5, 7, 7$ wins for $m=50,100,200$ respectively.  Again we see that \texttt{PenVF} performs poorly with \texttt{abalone}, \linebreak \texttt{winequality-red} and \texttt{winequality-white}.

\setlength{\tabcolsep}{1.5pt}

\begin{table*}[ht]
\centering
\small
\begin{tabular}{l | l l l | l l l | l l l}

\hline
& \multicolumn{3}{|c|}{$m=50$} & \multicolumn{3}{|c|}{$m=100$} & \multicolumn{3}{|c}{$m=200$} \\
 & VFCV & PenVF+ & PenVF & VFCV & PenVF+ & PenVF  & VFCV & PenVF+ & PenVF  \\
\hline 
& \multicolumn{9}{c}{SVR} \\
\hline
ab & .544 (.031) & .549 (.033) & .624 (.036) & .514 (.017) & .519 (.027) & .603 (.026) & .497 (.013) & .501 (.019) & .584 (.012)\\
ad & .471 (.037) & .471 (.031) & .473 (.036) & .409 (.020) & \textbf{.401} (.022) & \textbf{.404} (.023) & .330 (.021) & \textbf{.320} (.016) & \textbf{.320} (.013)\\
ca & .236 (.052) & .227 (.055) & \textbf{.220} (.060) & .172 (.051) & .165 (.048) & .165 (.048) & .130 (.026) & .127 (.021) & .128 (.021)\\
cc & .472 (.046) & .478 (.050) & .485 (.050) & .413 (.025) & .415 (.031) & .430 (.036) & .366 (.021) & \textbf{.350} (.021) & .366 (.020)\\
pm & .272 (.030) & .274 (.030) & .282 (.032) & .237 (.012) & .240 (.017) & .254 (.024) & .217 (.009) & .215 (.010) & .219 (.010)\\
pt & .116 (.024) & .116 (.025) & .114 (.024) & .092 (.012) & .092 (.013) & .090 (.014) & .080 (.009) & .078 (.009) & .078 (.010)\\
pd & .814 (.032) & .813 (.037) & \textbf{.806} (.026) & .796 (.021) & .797 (.028) & .793 (.015) & .778 (.014) & .783 (.018) & .779 (.015)\\
sl & .423 (.037) & .427 (.043) & .415 (.037) & .341 (.023) & .346 (.022) & \textbf{.334} (.020) & .290 (.012) & .292 (.018) & \textbf{.279} (.009)\\
wr & .712 (.064) & .709 (.062) & .745 (.057) & .660 (.034) & .667 (.041) & .710 (.048) & .637 (.021) & .643 (.030) & .704 (.037)\\
ww & .728 (.028) & .734 (.039) & .752 (.034) & .704 (.025) & .710 (.035) & .731 (.034) & .680 (.025) & .679 (.023) & .707 (.031)\\
\hline 
& \multicolumn{9}{c}{CART} \\
\hline
ab & .699 (.057) & .696 (.065) & .713 (.065) & .665 (.044) & .667 (.049) & .691 (.052) & .633 (.028) & .632 (.040) & .661 (.036)\\
ad & .750 (.063) & \textbf{.725} (.066) & \textbf{.720} (.072) & .637 (.055) & \textbf{.616} (.048) & \textbf{.624} (.040) & .567 (.036) & \textbf{.556} (.027) & .571 (.030)\\
ca & .330 (.116) & \textbf{.305} (.100) & \textbf{.303} (.099) & .220 (.052) & .215 (.051) & .215 (.050) & .186 (.031) & .180 (.023) & .182 (.023)\\
cc & .681 (.084) & \textbf{.642} (.068) & \textbf{.624} (.056) & .569 (.069) & \textbf{.535} (.051) & \textbf{.526} (.045) & .459 (.037) & \textbf{.444} (.030) & \textbf{.438} (.029)\\
pm & .348 (.055) & \textbf{.331} (.038) & \textbf{.323} (.036) & .282 (.037) & \textbf{.268} (.028) & \textbf{.262} (.027) & .210 (.025) & \textbf{.204} (.022) & \textbf{.202} (.021)\\
pt & .268 (.160) & .259 (.161) & .257 (.161) & .213 (.077) & \textbf{.198} (.044) & \textbf{.198} (.045) & .161 (.020) & .158 (.020) & .157 (.020)\\
pd & .812 (.036) & .821 (.054) & 1.039 (.089) & .796 (.014) & .794 (.048) & .947 (.104) & .751 (.051) & \textbf{.694} (.057) & .808 (.077)\\
sl & .679 (.128) & \textbf{.604} (.105) & \textbf{.594} (.102) & .484 (.088) & \textbf{.453} (.071) & \textbf{.444} (.069) & .362 (.048) & \textbf{.344} (.032) & \textbf{.340} (.030)\\
wr & .812 (.067) & .797 (.065) & .799 (.072) & .776 (.059) & \textbf{.756} (.061) & .765 (.066) & .738 (.045) & \textbf{.721} (.044) & .730 (.048)\\
ww & .789 (.052) & .786 (.065) & .807 (.072) & .771 (.040) & \textbf{.762} (.041) & .782 (.059) & .760 (.035) & \textbf{.744} (.038) & .768 (.053)\\
\hline
\end{tabular}
\caption{Error rates (with standard deviations in parentheses) for cross validation with the SVR (top) and CART (bottom) and penalisation for $V=2$, $\alpha=1.0$. Statistically significant improvements over VFCF are in bold. With the SVR, \texttt{PenVF+} is generally comparable to VFCV and \texttt{PenVF} is more variable. With CART, the penalisation methods both improve over VFCV a number of times. }
\label{tab:allDatasets} 
\end{table*}

\setlength{\tabcolsep}{5pt}

\subsection{Paired $t$-test Comparison with VFCV} 

Table \ref{tab:winsCounts} shows the results of the paired $t$-tests to compare \texttt{PenVF} with $\alpha=1.0$ and \texttt{PenVF+} with VFCV. Consider first the SVR results. Here we see that as one might expect, there are few statistically significant differences between $10$-fold CV and \texttt{PenVF}. Indeed, results indicate that as $V$ increases the penalisation methods and VFCV become more similar. In particular we see that \texttt{PenVF+} is identical to VFCV in all but one or two cases, with the main exception being 5 draws with $m=200$ and $V=2$, in which there are 2 improvements and 3 losses (\texttt{abalone}, \texttt{pumadyn-32nh} and \texttt{winequality-red}). \texttt{PenVF} fares worse against VFCV: we see more wins but at the same time more losses. Our later analysis will shed light on why this is the case. We also compared the ``ideal'' model, in which the test set is used during model selection, with VFCV and found that one can generally gain improvements except in the case of \texttt{slice-
loc}, which has a large number of features. In every case the ideal model selector can improve over 2-fold CV. 

The CART results at the bottom of Table \ref{tab:winsCounts} show that for $V > 4$ penalisation is identical to VFCV. Clearly the bias with low values of $V$ in conjunction with VFCV is more prominent in this case. We have already examined the $V=2$ case and with $V=4$ we see improvements in one case for \texttt{PenVF+} for $m=50, 200$.  With the ideal errors we see that with more of the datasets compared to SVR, performance cannot be improved by using the test realisation and furthermore as $m$ increases improvements over VFCV are increasingly difficult except for the $2$-fold case.   

\begin{table*}[ht]
\centering
\begin{tabular}{l |  l l l l l l | l l l l l l | l l l l l l }
\hline
 & 2 & 4 & 6 & 8 & 10 & 12 & 2 & 4 & 6 & 8 & 10 & 12  & 2 & 4 & 6 & 8 & 10 & 12  \\ 
\hline
& \multicolumn{18}{c}{SVR} \\
\hline 
PenVF+  & 0 & 0 & 0 & 1 & 0 & 0 & 10 & 10 & 10 & 9 & 10 & 10 & 0 & 0 & 0 & 0 & 0 & 0\\
PenVF  & 5 & 4 & 3 & 1 & 1 & 0 & 3 & 5 & 7 & 9 & 9 & 10 & 2 & 1 & 0 & 0 & 0 & 0\\
Ideal & 0 & 0 & 0 & 0 & 0 & 0 & 0 & 1 & 1 & 1 & 1 & 1 & 10 & 9 & 9 & 9 & 9 & 9\\
\hline
PenVF+  & 0 & 1 & 0 & 0 & 0 & 0 & 9 & 9 & 10 & 10 & 10 & 10 & 1 & 0 & 0 & 0 & 0 & 0\\
PenVF  & 5 & 5 & 3 & 3 & 2 & 1 & 3 & 3 & 6 & 6 & 8 & 9 & 2 & 2 & 1 & 1 & 0 & 0\\
Ideal  & 0 & 0 & 0 & 0 & 0 & 0 & 0 & 1 & 1 & 1 & 1 & 1 & 10 & 9 & 9 & 9 & 9 & 9\\
\hline
PenVF+ & 3 & 1 & 0 & 1 & 0 & 0 & 5 & 9 & 10 & 9 & 10 & 10 & 2 & 0 & 0 & 0 & 0 & 0\\
PenVF  & 3 & 4 & 3 & 3 & 2 & 1 & 5 & 5 & 6 & 6 & 7 & 8 & 2 & 1 & 1 & 1 & 1 & 1\\
Ideal  & 0 & 0 & 0 & 0 & 0 & 0 & 0 & 1 & 1 & 1 & 1 & 1 & 10 & 9 & 9 & 9 & 9 & 9\\
\hline
& \multicolumn{18}{c}{CART} \\
\hline
PenVF+ & 0 & 0 & 0 & 0 & 0 & 0 & 5 & 9 & 10 & 10 & 10 & 10 & 5 & 1 & 0 & 0 & 0 & 0\\
PenVF & 2 & 1 & 0 & 0 & 0 & 0 & 3 & 6 & 10 & 10 & 10 & 10 & 5 & 3 & 0 & 0 & 0 & 0\\
Ideal & 0 & 0 & 0 & 0 & 0 & 0 & 1 & 2 & 2 & 2 & 2 & 2 & 9 & 8 & 8 & 8 & 8 & 8\\
\hline
PenVF+  & 0 & 0 & 0 & 0 & 0 & 0 & 3 & 10 & 10 & 10 & 10 & 10 & 7 & 0 & 0 & 0 & 0 & 0\\
PenVF  & 2 & 3 & 0 & 0 & 0 & 0 & 3 & 5 & 10 & 10 & 10 & 10 & 5 & 2 & 0 & 0 & 0 & 0\\
Ideal & 0 & 0 & 0 & 0 & 0 & 0 & 1 & 2 & 4 & 3 & 4 & 3 & 9 & 8 & 6 & 7 & 6 & 7\\
\hline
PenVF+  & 0 & 0 & 0 & 0 & 0 & 0 & 3 & 9 & 10 & 10 & 10 & 10 & 7 & 1 & 0 & 0 & 0 & 0\\
PenVF & 2 & 2 & 0 & 0 & 0 & 0 & 5 & 8 & 10 & 10 & 10 & 10 & 3 & 0 & 0 & 0 & 0 & 0\\
Ideal  & 0 & 0 & 0 & 0 & 0 & 0 & 0 & 4 & 4 & 4 & 5 & 4 & 10 & 6 & 6 & 6 & 5 & 6\\
\hline
\end{tabular}
\caption{Number of statistically significant losses (left block), draws (middle block) and wins (right block) against standard errors for CV and across different numbers of folds using $\alpha=1.0$. The sample size $m$ is 50 (top), 100 (middle), and 200 (bottom). As $V$ increases the penalisation methods become more similar to VFCV. In particular for CART, when $V > 4$ penalisation is identical to VFCV. }
\label{tab:winsCounts}
\end{table*}

\subsection{Optimal Penalisation Constant is Dataset Dependent} 

\begin{figure}[ht]
\centering
\subfigure[SVR, \texttt{abalone} (top) and \texttt{slice-loc} (bottom)]{\includegraphics[width=0.80\linewidth]{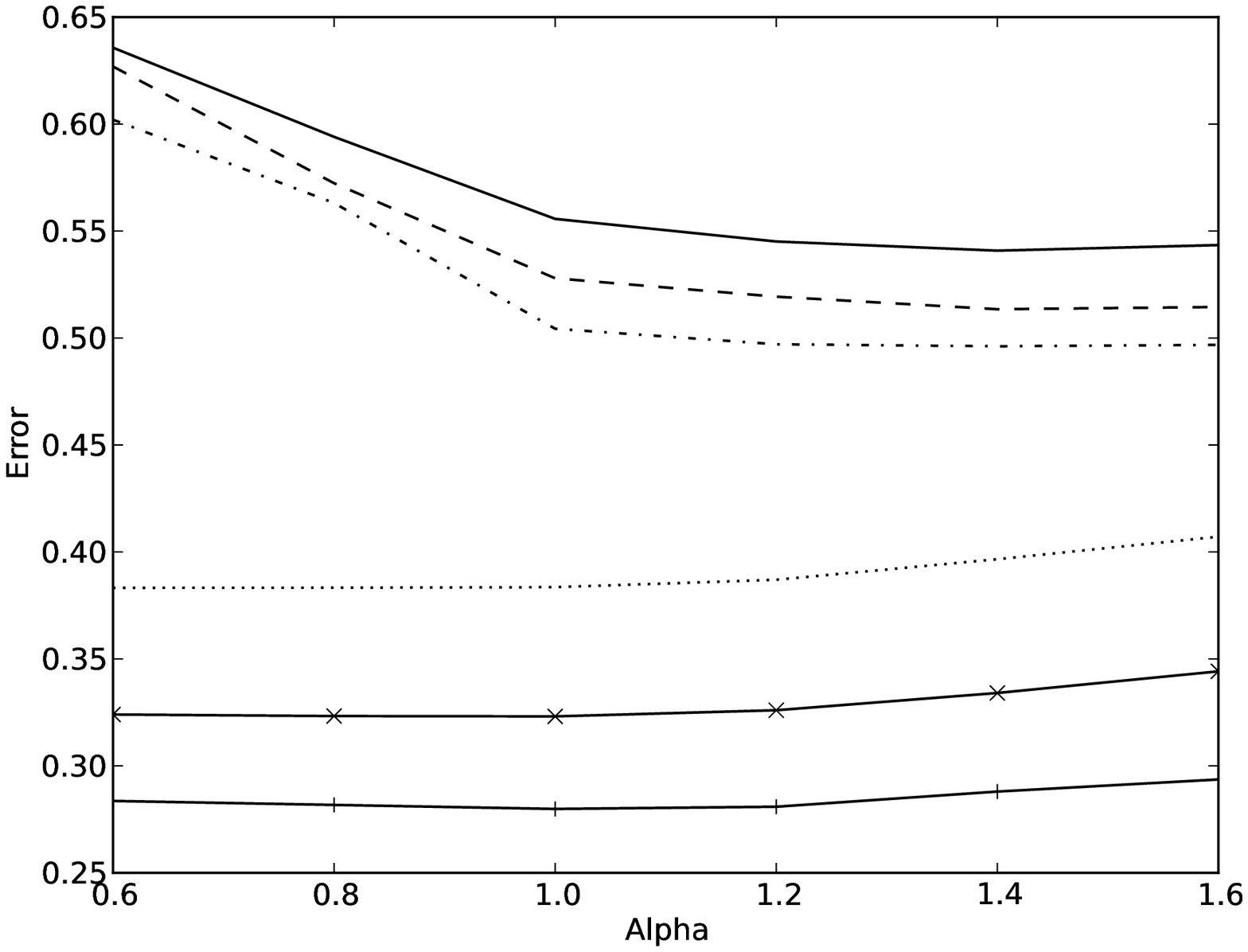}}
\subfigure[CART, \texttt{winequality-white} (top) and \texttt{slice-loc} (bottom)]{\includegraphics[width=0.80\linewidth]{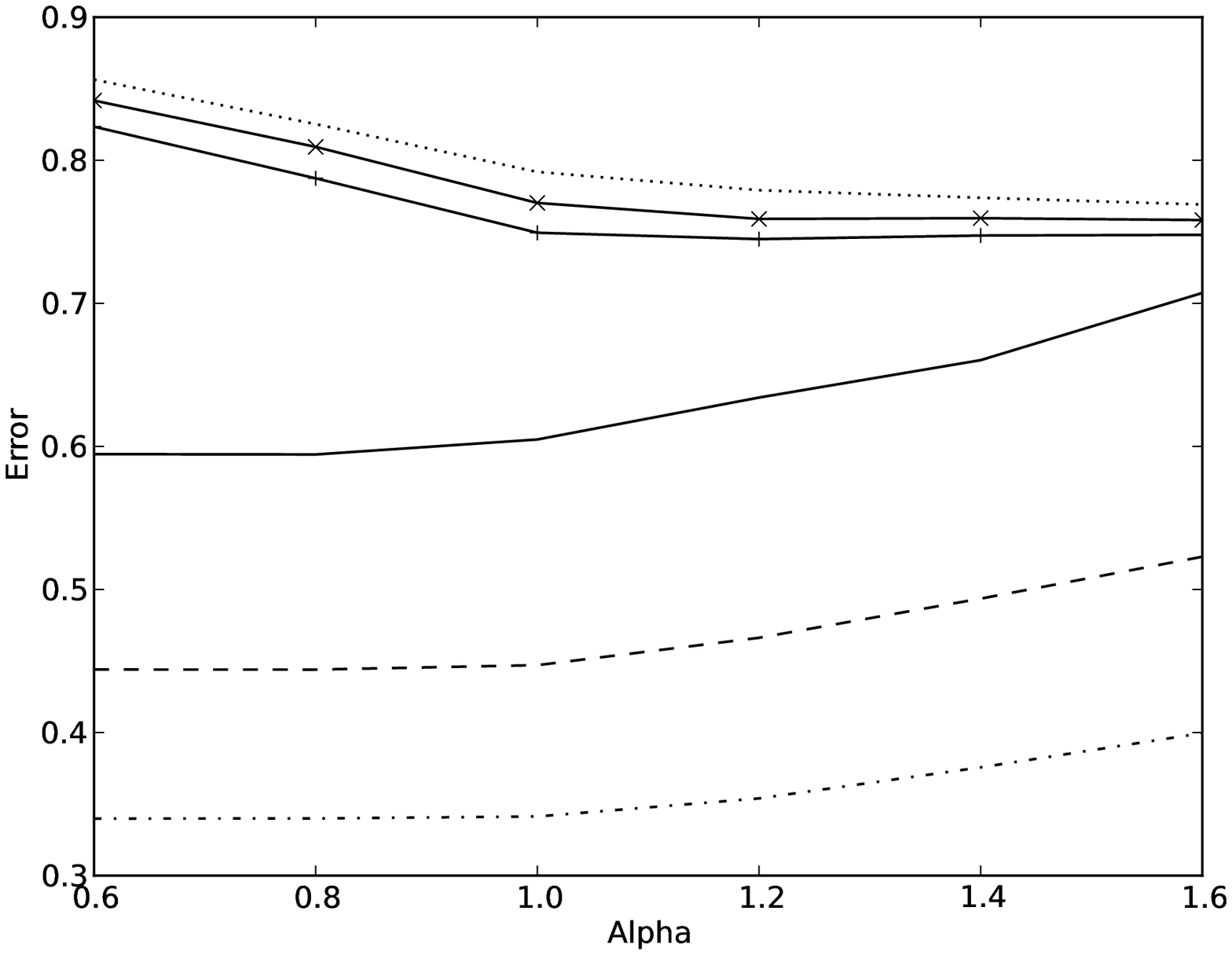}} 
\caption{\label{fig:alpha} The variation in error with $\alpha$ for some sample datasets with $V=10$. The plots are ordered from top to bottom: $m=50, 100, 200$, for example for \texttt{winequality-white} and CART the respective curves are dotted, solid with crosses and solid with pluses. With \texttt{abalone} and the SVR, and \texttt{winequality-white} with CART overpenalisation improves results, however it makes them worse with \texttt{slice-loc} for both the SVR and CART. } 
\end{figure}

To discover the effect of overpenalisation, observe Figure \ref{fig:alpha} which shows the errors on 2 datasets as $\alpha$ varies when $V=10$. The effect of $\alpha$ is clearly dataset dependent: on \texttt{abalone} observe that the error tends to decrease with the SVR with more penalisation, and hence a slight amount of overpenalisation ($\alpha = 1.2$) is recommended. In contrast, overpenalisation increases the error with \texttt{slice-loc}. In total, 5 of the datasets benefited from overpenalisation and 3 improved with underpenalisation. With \texttt{pumadyn} and \texttt{parkinsons-total}, a value of $\alpha=1.0$ as predicted by the theory gave optimal results. Results were similar with CART except that 7 of the datasets benefit from underpenalisation. Notice that as sample size $m$ increases the choice of $\alpha$ becomes less important since estimates of error for each model are more reliable and the penalty terms become small. We noticed that with CART, VFCV consistently underestimates the tree size 
whereas \texttt{PenVF} chooses larger sizes in general and this can be an advantage or disadvantage depending on the variation of error with $t$. Also observed is that as expected VFCV provides a pessimistic error compared to the ideal case and \texttt{PenVF} is generally more accurate than VFCV, however, as in model selection since we pick the model with the lowest error, this does not always translate into the best predictor. 

\subsection{On the Estimation of the Ideal Penalty}

\begin{figure}[ht]
\centering
\subfigure[$m=50$]{\includegraphics[width=0.80\linewidth]{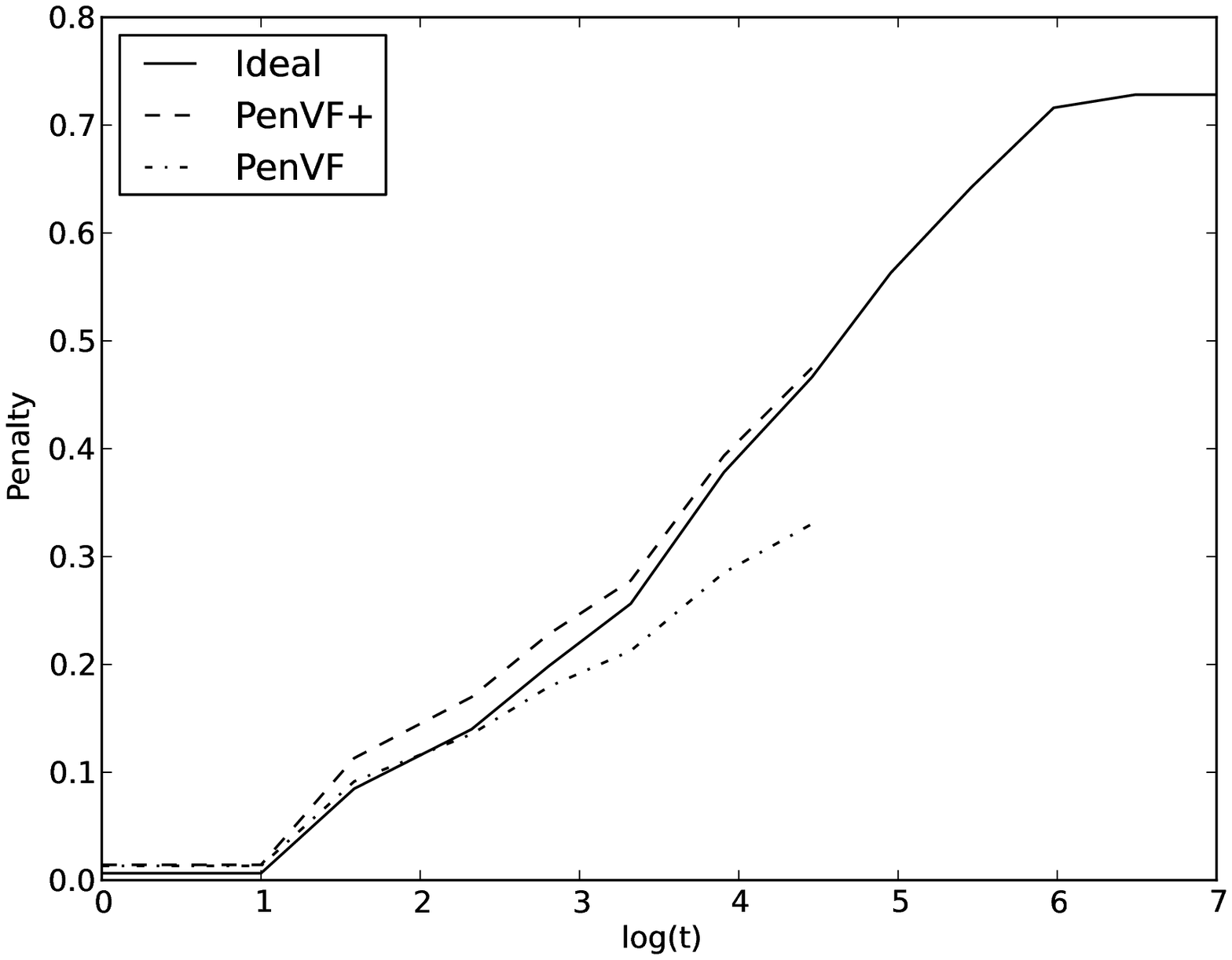}}
\subfigure[$m=100$]{\includegraphics[width=0.80\linewidth]{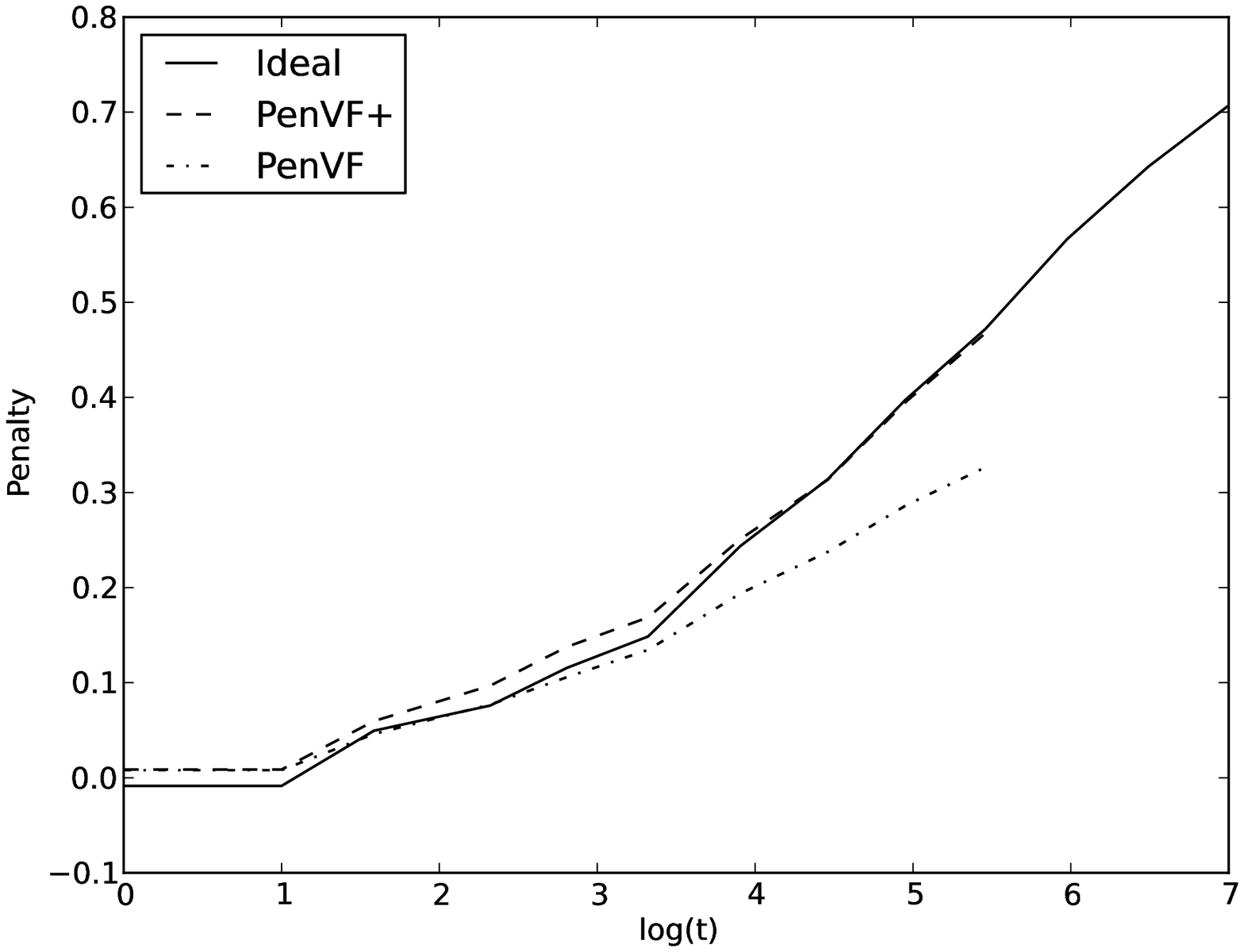}}
\caption{\label{fig:penalties} The variation in penalty for CART in the ``ideal'' case relative to \texttt{PenVF} $\alpha=1.0$, and PenVF+ with $V=2$ and \texttt{abalone}. \texttt{PenVF+} estimates the ideal error well across a range of $t$'s, whereas \texttt{PenVF} underestimates it for large $t$. } 
\end{figure}

Next we study the approximated penalty and how it differs from the ``ideal'' penalty in the case of CART with $V=2$, see Figure \ref{fig:penalties}. Notice that the curves for \texttt{PenVF} and \texttt{PenVF+} are shorter than the ones for the ``ideal'' penalty since only half the examples are used for training, limiting the tree size. The \texttt{PenVF} method diverges from the ideal case when we grow large trees, but is close to the ideal case for small trees. This change occurs with relatively small trees: size 4 with $m=50$ and  size 11 with $m=100$. This pattern was observed with most of the datasets. When we looked at a greater number of folds, \texttt{PenVF} was close to the ideal penalty. In contrast, \texttt{PenVF+} does not diverge as the tree size increases, however it seems to slightly overestimate the penalty. 

\begin{figure}[ht]
\centering
\includegraphics[width=0.80\linewidth]{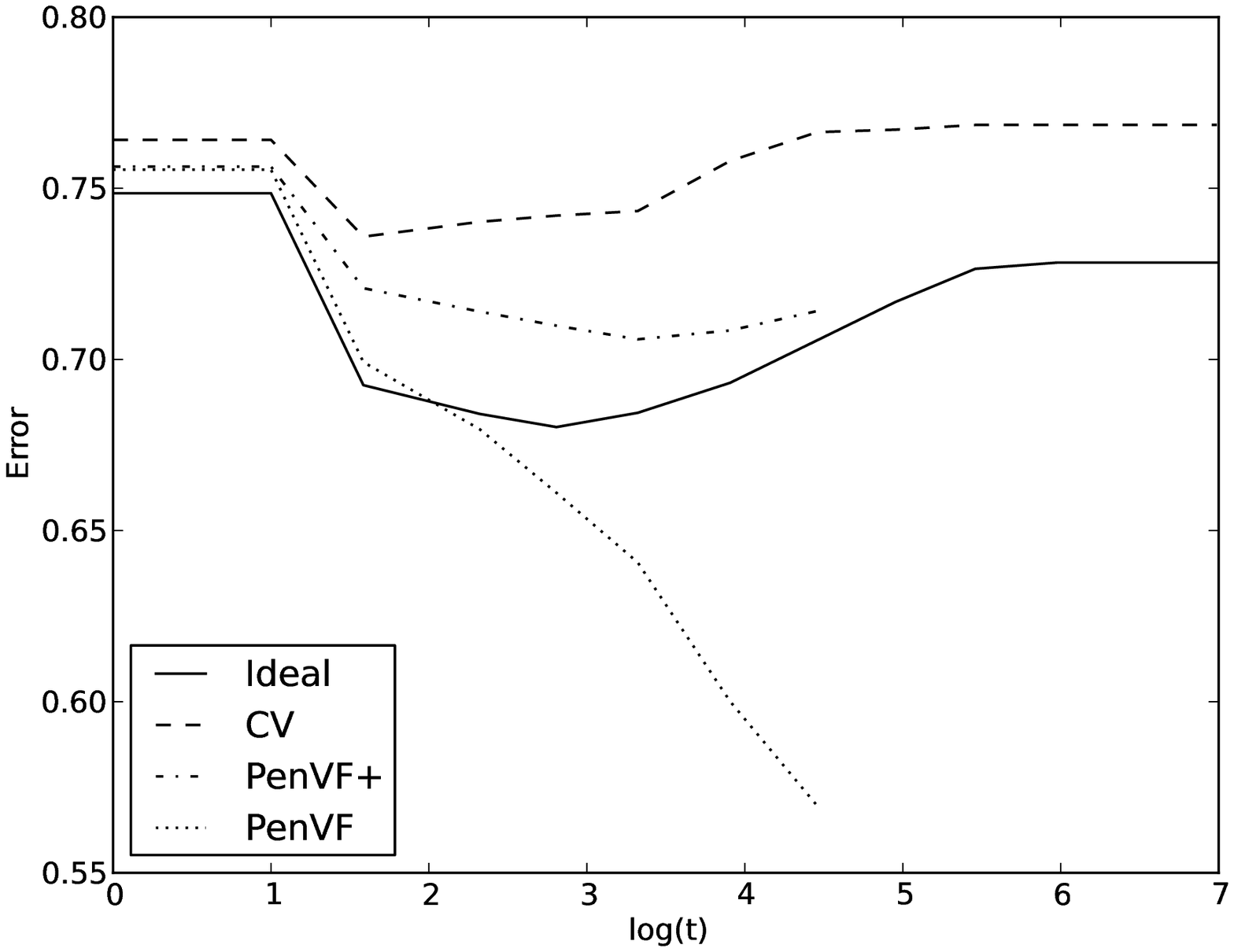}
\caption{\label{fig:CARTErrors} The error for CART in the ``ideal'' case relative to \texttt{PenVF} $\alpha=1.0$, and PenVF+ with $V=2$ and \texttt{abalone}, $m=50$. \texttt{PenVF} provides a poor estimation of the error for large values of $t$. \texttt{PenVF+} gives better error estimates but results in the selection of larger trees than VFCV. } 
\end{figure}

The question remains about which cases \texttt{PenVF} improves over VFCV for low values of $V$ for CART. Figure \ref{fig:CARTErrors} demonstrates that the error estimation for \texttt{PenVF} is generally optimistic as model size increases. With \texttt{abalone} for example the optimal tree was of size 7 nodes, however \texttt{PenVF} chooses one of size 22. In contrast, on some datasets large trees did not overfit the test set and hence in these cases \texttt{PenVF} can perform better than VFCV. This also sheds some light on Figure \ref{fig:alpha} where we see that overpenalisation helps in some cases but not in others. \texttt{PenVF+} provides the best estimate of the error in general, however it also results in a larger tree size that the ideal case.

\subsection{Extended Setup}

In this final set of experiments we explore further the distinction between penalisation and VFCV with CART by considering $m=500$ and using the same set of folds as in the original setup, see Table \ref{tab:winsCountsExtCART}. The interesting figures in this case are those corresponding to 2 and 4 folds in which we see that \texttt{PenVF+} wins for 8 and 2 datasets respectively. In fact, since in the ideal case one can only win 8 times with $V=2$ this certainly demonstrates the effectiveness of \texttt{PenVF+} and bias in VFCV in this case.

\begin{table*}
\centering
\begin{tabular}{l |  l l l l l l | l l l l l l | l l l l l l }
\hline
 & 2 & 4 & 6 & 8 & 10 & 12 & 2 & 4 & 6 & 8 & 10 & 12  & 2 & 4 & 6 & 8 & 10 & 12  \\ 
\hline
PenVF+  & 0 & 0 & 0 & 0 & 0 & 0 & 2 & 8 & 10 & 10 & 10 & 10 & 8 & 2 & 0 & 0 & 0 & 0\\
PenVF & 2 & 0 & 0 & 0 & 0 & 0 & 3 & 7 & 9 & 10 & 10 & 10 & 5 & 3 & 1 & 0 & 0 & 0\\
Ideal & 0 & 0 & 0 & 0 & 0 & 0 & 2 & 3 & 4 & 4 & 4 & 4 & 8 & 7 & 6 & 6 & 6 & 6\\
\hline
\end{tabular}
\caption{Wins, draws and losses for CART using $m=500$. Note that \texttt{PenVF+} wins 8 times for $V=2$ and 2 times for $V=4$.}
\label{tab:winsCountsExtCART}
\end{table*}

\subsection{Key Points} 

Our experimental analysis has painted a detailed picture of penalisation versus cross validation for model selection. The bias in VFCV is evident with small values of $V$ and small training sets, and we observed that as $V$ and the training set sizes increase the model selection methods become more similar. With the SVR, \texttt{PenVF} makes a number of loses relative to VFCV and these losses are nearly all corrected with our modified penalisation \texttt{PenVF+}. Penalisation is more effective in general with CART: when $V>4$ both \texttt{PenVF} and \texttt{PenVF+} are not statistically significantly different to VFCV, and for $V=2$, \texttt{PenVF+} is at least as good as VFCV or improves over it in nearly every case, winning $5, 7, 7$ times for $m=50, 100, 200$ examples. In contrast \texttt{PenVF} is more variable in comparison to VFCV and one reason for this is that it underestimates the penalty to a large degree with large models. On some datasets larger trees did not increase the error and hence in 
these cases \texttt{PenVF} performs well. In general \texttt{PenVF+} provides a much better approximation of the ideal penalty compared to \texttt{PenVF}. The most striking results were with CART and $V=2$ in which we saw that \texttt{PenVF+} improves over VFCV in 8 out of 10 cases with $m=500$ examples. 

\section{Conclusions\label{section_conclu}}

Model selection is a critical part of machine learning as it can dramatically affect generalisation performance. In practice, cross validation over a grid of parameter values is often used, and it has been shown to be very effective in a variety of cases. We studied $V$-fold penalisation which is a general purpose penalisation procedure that aims at improving on VFCV by correcting its bias and is proved in \cite{Arl:2008a} to be asymptotically optimal in a histogram regression setting. $V$-fold penalisation is simple to implement and the penalised error can be computed using the same predictions as cross validation and hence at negligible additional computational cost. Furthermore, we propose an improvement of penalisation, called \texttt{PenVF+}, which takes into account learning rates in order to correct under-penalisation with large models. 

We conducted an extensive empirical investigation into VFCV and V-fold penalisation over a collection of 10 well known benchmark datasets using an SVR with the RBF kernel and CART.  With low values of $V$, penalisation can provide an advantage over VFCV but this advantage rapidly diminishes as $V$ increases. Furthermore, in some cases penalisation fared worse than cross validation. When we compare the penalty with the ``ideal'' penalty, we observed that \texttt{PenVF} underestimates the penalty with large models and \texttt{PenVF+} improves penalty estimation in these cases. Hence, there is no fixed overpenalisation constant even for a particular dataset, but rather the penalisation should vary with model complexity as with \texttt{PenVF+}. 

\bibliographystyle{plain}  
\bibliography{references}

\end{document}